\documentclass[journal]{IEEEtran}
\usepackage{cite}

\ifCLASSINFOpdf
  \usepackage[pdftex]{graphicx}
  \graphicspath{{../pdf/}{../jpeg/}}
  \DeclareGraphicsExtensions{.pdf,.jpeg,.png}
\else
  \usepackage[dvips]{graphicx}
  \graphicspath{{../eps/}}
  \DeclareGraphicsExtensions{.eps}
\fi
\usepackage{amsmath}
\usepackage{algorithmic}

\usepackage{array}

\ifCLASSOPTIONcompsoc
 \usepackage[caption=false,font=normalsize,labelfont=sf,textfont=sf]{subfig}
\else
 \usepackage[caption=false,font=footnotesize]{subfig}
\fi
\usepackage{url}

\usepackage[linesnumbered,boxed,ruled,commentsnumbered]{algorithm2e}
\usepackage{booktabs}
\usepackage{multirow}
\usepackage{colortbl}
\definecolor{mygray}{gray}{.9}

\newcommand{\eg}{\textit{e}.\textit{g}.}
\newcommand{\ie}{\textit{i}.\textit{e}.}
\usepackage{color}
\newcommand{\blue}[1]{\textcolor{black}{#1}}
\begin{document}
%
\title{Deep Learning for Weakly-Supervised Object Detection and Object Localization: A Survey}

%
%
%


\author{Feifei~Shao,
        Long~Chen,
        Jian~Shao~\IEEEmembership{Member,~IEEE,}
        Wei~Ji,
        Shaoning~Xiao,
        Lu~Ye,
        Yueting~Zhuang~\IEEEmembership{Member,~IEEE,}
        Jun~Xiao~\IEEEmembership{Member,~IEEE,}
\thanks{Feifei Shao, Jian Shao, Shaoning Xiao, Yueting Zhuang, Jun Xiao are with Zhejiang University, Hangzhou, 310027, China. Emails: \{sff, jshao, shaoningx, yzhuang, junx\}@zju.edu.cn}
\thanks{Long Chen is with Columbia University, New York, 10027, USA. This work was partially done when L. Chen was a Ph.D. student at Zhejiang University. Email: zjuchenlong@gmail.com (Corresponding author).}
\thanks{Wei Ji is with National University of Singapore, 117417, Singapore. Email: weiji0523@gmail.com}
\thanks{Lu Ye is with Zhejiang University of Science and Technology, Hangzhou, 310023, China. Email: yelue@zust.edu.cn}
}

\maketitle

\begin{abstract}
Weakly-Supervised Object Detection (WSOD) and Localization (WSOL), \ie, detecting multiple and single instances with bounding boxes in an image using image-level labels, are long-standing and challenging tasks in the CV community. With the success of deep neural networks in object detection, both WSOD and WSOL have received unprecedented attention. Hundreds of WSOD and WSOL methods and numerous techniques have been proposed in the deep learning era. To this end, in this paper, we consider WSOL is a sub-task of WSOD and provide a comprehensive survey of the recent achievements of WSOD. Specifically, we firstly describe the formulation and setting of the WSOD, including the background, challenges, basic framework. Meanwhile, we summarize and analyze all advanced techniques and training tricks for improving detection performance. Then, we introduce the widely-used datasets and evaluation metrics of WSOD. Lastly, we discuss the future directions of WSOD. We believe that these summaries can help pave a way for future research on WSOD and WSOL.
\end{abstract}


\ifCLASSOPTIONpeerreview
\begin{center} \bfseries EDICS Category: 3-BBND \end{center}
\fi
%
\IEEEpeerreviewmaketitle

\section{Introduction}
\IEEEPARstart{O}{bject} detection~\cite{zhao2019object, liu2020deep} is a fundamental and challenging task that aims to locate and classify object instances in an image. The object localization is to use a bounding box (an axis-aligned rectangle tightly bounding the object) to search for the spatial location and range of an object in an image as much as possible~\cite{everingham2010pascal, russakovsky2015imagenet}. Object classification is to assess the presence of objects from a given set of object classes in an image. As one of the most fundamental tasks in computer vision, object detection is an indispensable technique for many high-level applications, \eg, robot vision~\cite{james2019sim}, face recognition~\cite{hu2017finding}, image retrieval~\cite{bai2021unsupervised, chen2021feature}, augmented reality~\cite{tomei2019art2real}, autonomous driving~\cite{behl2017bounding}, change detection~\cite{liu2016deep} and so on. With the development of convolutional neural networks (CNNs) in visual recognition~\cite{simonyan2014very, szegedy2015going, he2016deep} and release of large scale dateset~\cite{russakovsky2015imagenet,lin2014microsoft}, today's state-of-the-art object detector can achieve near-perfect performance under fully-supervised setting, \ie, Fully-Supervised Object Detection (FSOD)~\cite{ren2015faster, liu2016ssd, redmon2016you, he2017mask, lin2017feature, law2018cornernet}. Unfortunately, these fully-supervised object detection methods suffer from two inevitable limitations: 1) The large-scale instance annotations are difficult to obtain and labor-intensive. 2) When labeling these data, they may introduce annotation noises inadvertently.

\begin{figure}[t]
  \begin{center}
      \includegraphics[width=0.85\linewidth]{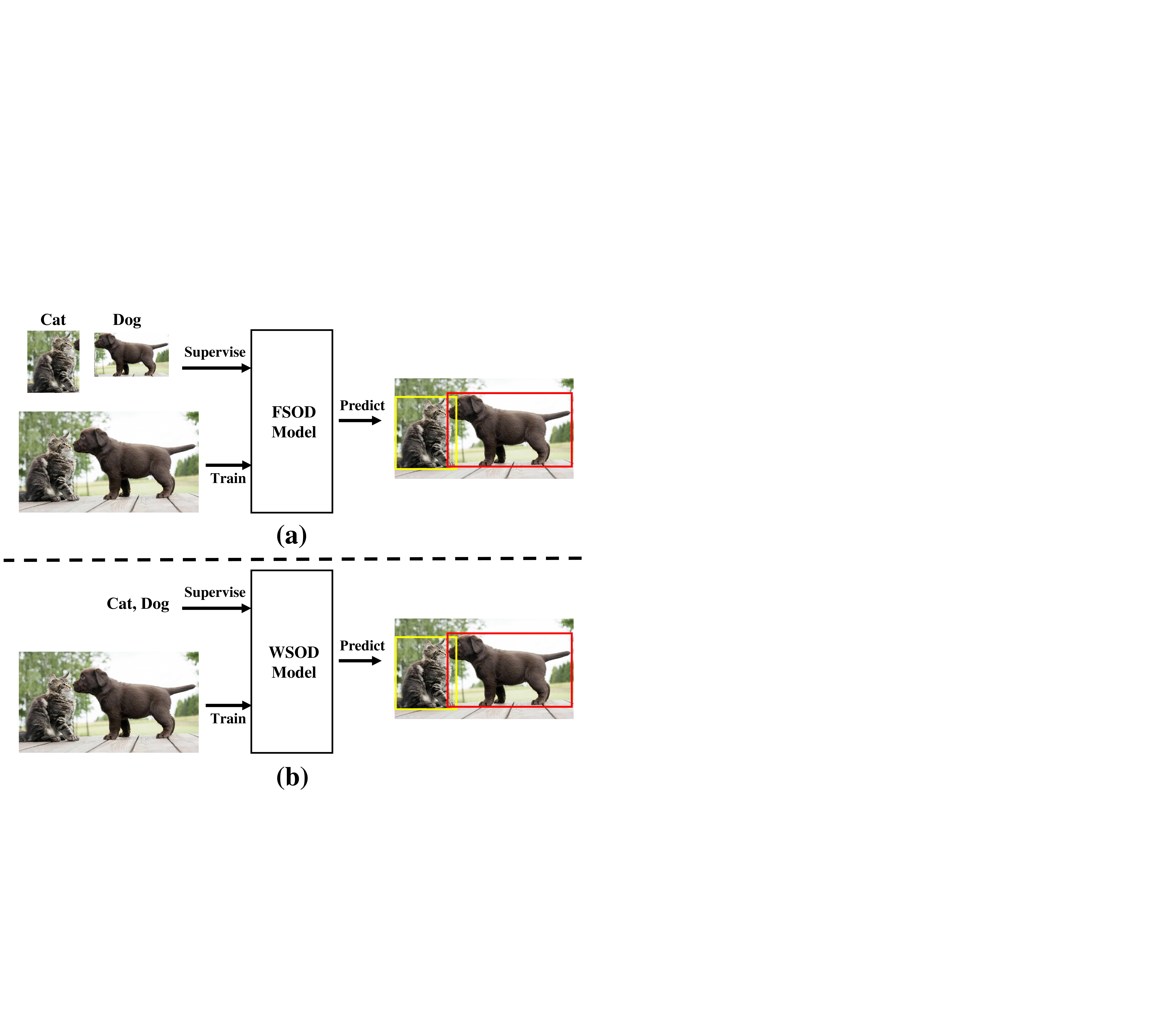}
  \end{center}
  \caption{(a) Fully-Supervised Object Detection (FSOD) uses the \emph{instance-level} annotations as supervision. (b) Weakly-Supervised Object Detection (WSOD) uses the \emph{image-level} annotations as supervision.}
  \label{fsod_vs_wsod}
\end{figure}

To avoid the mentioned problems, the community starts to solve object detection in a weakly-supervised setting, \ie, Weakly-Supervised Object Detection (WSOD). Different from the fully-supervised setting (cf. Fig.~\ref{fsod_vs_wsod} (a)), WSOD aims to detect instances with only image-level labels (\eg, categories of instances in the whole images). Meanwhile, WSOD can benefit from the large-scale datasets on the web, such as Facebook and Twitter. Another similar task is Weakly-Supervised Object Localization (WSOL), which only detects one instance in an image. Because WSOD and WSOL detect multiple instances and single instances respectively, we consider WSOL as a sub-task of WSOD. In the following paper, we use WSOD to represent both WSOD and WSOL.

\begin{figure*}[t]
  \begin{center}
      \includegraphics[width=0.9\linewidth]{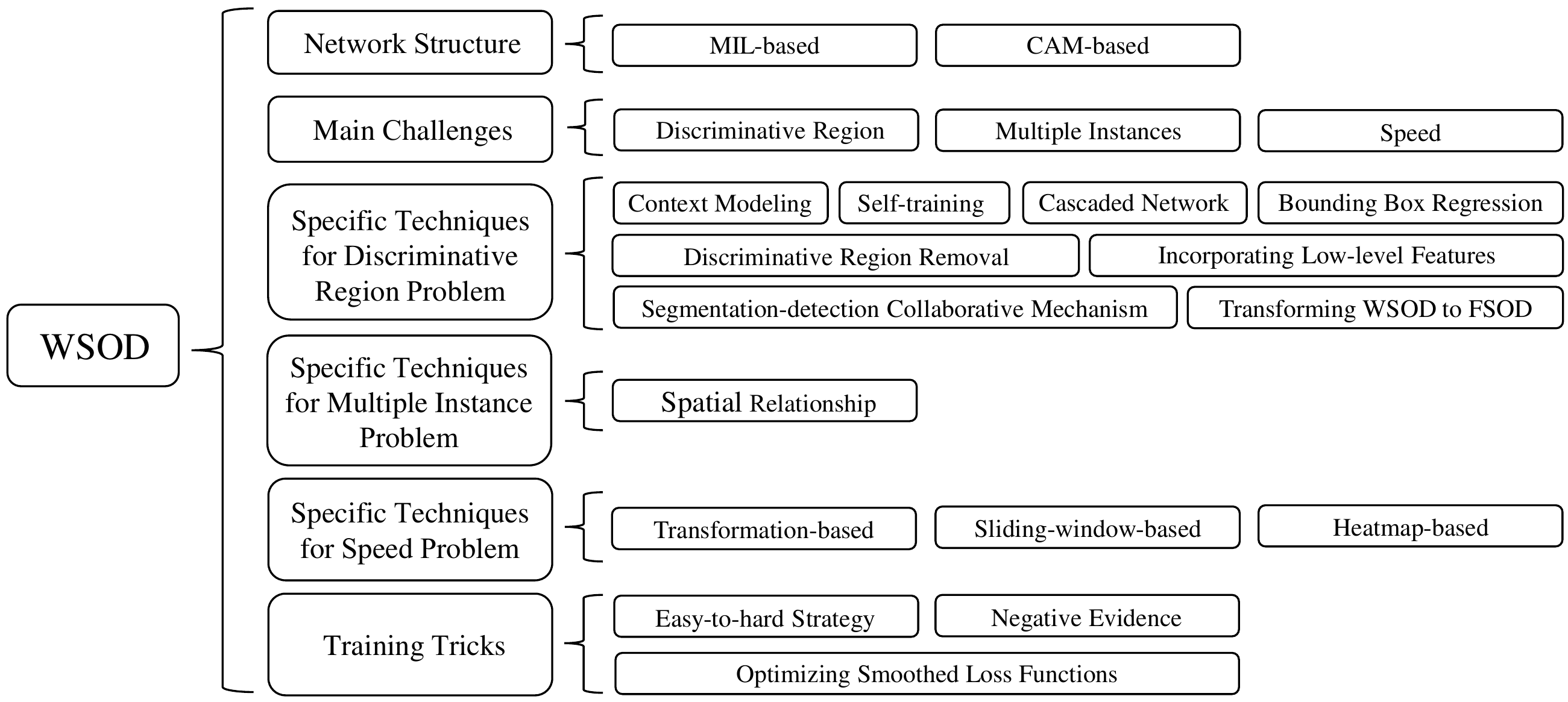}
  \end{center}
  \caption{The main content of this paper.}
  \label{sections_figure}
\end{figure*}

In this paper, we go over all typical WSOD methods and give a comprehensive survey (cf. Fig.~\ref{sections_figure}) of recent advances in WSOD. Since the number of papers on WSOD is breathtaking, we sincerely apologize to those authors whose research on WSOD and other related fields are not included in this survey. In Section~\ref{sec:wsod}, we introduce the background, main challenges, and basic framework. In Section~\ref{sec:milestones}, according to the development timeline of WSOD, we introduce several modern classical methods in detail. Then, in-depth analyses are provided towards the all advanced techniques and tricks for the main challenges. In Section~\ref{sec:datasets}, we demonstrate all prevailing benchmarks and standard evaluation metrics for WSOD. In Section~\ref{sec:directions}, we briefly discuss the future directions.


\section{WSOD}
\label{sec:wsod}
\subsection{A Problem Definition}

\blue{WSOD aims to classify and locate object instances using only image-level labels in the training phase. As shown in Fig.~\ref{fsod_vs_wsod} (b), given an image with cat and dog, WSOD not only classifies the cat and dog but also locates their location using bounding boxes. Different from FSOD that can use instance-level annotations in the training phase shown in Fig.~\ref{fsod_vs_wsod} (a), WSOD only accesses image-level labels. Because of this restriction, though hundreds of WSOD methods have been proposed, the performance gap between WSOD and FSOD is still large. For example, the mAP of state-of-the-art FSOD approach~\cite{singh2018sniper} and WSOD approach~\cite{ren2020instance} is 86.9\% and 54.9\% on PASCAL VOC 2007 dataset~\cite{everingham2007pascal}, respectively. Therefore, there are still many challenges in terms of the task of WSOD for researchers to solve, especially in the direction of improving the detection performance.}

\addtolength{\tabcolsep}{-3pt}
\begin{table*}[t]
  \centering
  \caption{A summary of the state-of-the-art WSOD methods. For the proposals, SS represents selective search, EB represents edge boxes, and SW represents sliding window. The Challenges denotes the main contributions of corresponding papers.}
      \label{table08}
      \begin{tabular}{lcccccccc}
      \toprule
      \multirow{2}*{Approach}&
      \multirow{2}*{Year} & 
      \multirow{2}*{Proposals} & 
      \multicolumn{2}{c}{Network}& 
      \multicolumn{3}{c}{Challenges} &
      \multirow{2}*{Code on Github} \\
      
      \cmidrule (r){4-5} \cmidrule (r){6-8}
      &&&MIL-based&CAM-based&Discriminative Region&Multiple Instances&Speed&\\
      \midrule
      \rowcolor{mygray}
      WSDDN~\cite{bilen2016weakly}&CVPR2016&EB&$\surd$&&&&&hbilen/WSDDN\\
      CAM~\cite{zhou2016learning}&CVPR2016&Heatmap&&$\surd$&&&$\surd$&zhoubolei/CAM\\
      \rowcolor{mygray}
      WSLPDA~\cite{li2016weakly}&CVPR2016&EB&$\surd$&&$\surd$&&&jbhuang0604/WSL\\
      WELDON~\cite{durand2016weldon}&CVPR2016&SW&$\surd$&&$\surd$&&$\surd$&\\
      \rowcolor{mygray}
      ContextLocNet~\cite{kantorov2016contextlocnet}&ECCV2016&SS&$\surd$&&$\surd$&&&vadimkantorov/contextlocnet\\
      Grad-CAM~\cite{selvaraju2017grad}&ICCV2017&Heatmap&&$\surd$&$\surd$&&$\surd$&ramprs/grad-cam\\
      \rowcolor{mygray}
      OICR~\cite{tang2017multiple}&CVPR2017&SS&$\surd$&&$\surd$&&&ppengtang/oicr\\
      WCCN~\cite{diba2017weakly}&CVPR2017&EB&$\surd$&&$\surd$&&&\\
      \rowcolor{mygray}
      ST-WSL~\cite{jie2017deep}&CVPR2017&EB&$\surd$&&$\surd$&$\surd$&&\\
      WILDCAT~\cite{durand2017wildcat}&CVPR2017&Heatmap&&$\surd$&$\surd$&&$\surd$&durandtibo/wildcat.pytorch\\
      \rowcolor{mygray}
      SPN~\cite{zhu2017soft}&ICCV2017&SW&$\surd$&&&&$\surd$&ZhouYanzhao/SPN\\
      TP-WSL~\cite{kim2017two}&ICCV2017&Heatmap&&$\surd$&$\surd$&&$\surd$&\\
      \rowcolor{mygray}
      PCL~\cite{tang2018pcl}&TPAMI2018&SS&$\surd$&&$\surd$&$\surd$&&ppengtang/pcl.pytorch\\
      GAL-fWSD~\cite{shen2018generative}&CVPR2018&EB&$\surd$&&&&$\surd$&\\
      \rowcolor{mygray}
      W2F~\cite{zhang2018w2f}&CVPR2018&SS&$\surd$&&$\surd$&$\surd$&$\surd$&\\
      ACoL~\cite{zhang2018adversarial}&CVPR2018&Heatmap&&$\surd$&$\surd$&&$\surd$&xiaomengyc/ACoL\\
      \rowcolor{mygray}
      ZLDN~\cite{zhang2018zigzag}&CVPR2018&EB&$\surd$&&$\surd$&&&\\
      TS$^2$C~\cite{wei2018ts2c}&ECCV2018&SS&$\surd$&&$\surd$&&&\\
      \rowcolor{mygray}
      SPG~\cite{zhang2018self}&ECCV2018&Heatmap&&$\surd$&&&$\surd$&xiaomengyc/SPG\\
      WSRPN~\cite{tang2018weakly}&ECCV2018&EB&$\surd$&&&&&\\
      \rowcolor{mygray}
      C-MIL~\cite{wan2019c}&CVPR2019&SS&$\surd$&&&&&WanFang13/C-MIL\\
      WS-JDS~\cite{shen2019cyclic}&CVPR2019&EB&$\surd$&&$\surd$&&&shenyunhang/WS-JDS\\
      \rowcolor{mygray}
      ADL~\cite{choe2019attention}&CVPR2019&Heatmap&&$\surd$&&&$\surd$&junsukchoe/ADL\\
      Pred NET~\cite{arun2019dissimilarity}&CVPR2019&SS&$\surd$&&&&&\\
      \rowcolor{mygray}
      WSOD2~\cite{zeng2019wsod2}&ICCV2019&SS&$\surd$&&$\surd$&&&researchmm/WSOD2\\
      OAILWSD~\cite{kosugi2019object}&ICCV2019&SS&$\surd$&&$\surd$&&&\\
      \rowcolor{mygray}
      TPWSD~\cite{yang2019towards}&ICCV2019&SS&$\surd$&&$\surd$&&&\\
      SDCN~\cite{li2019weakly}&ICCV2019&SS&$\surd$&&$\surd$&&&\\
      \rowcolor{mygray}
      C-MIDN~\cite{gao2019c}&ICCV2019&SS&$\surd$&&$\surd$&&&\\
      DANet~\cite{xue2019danet}&ICCV2019&Heatmap&&$\surd$&&&$\surd$&xuehaolan/DANet\\
      \rowcolor{mygray}
      NL-CCAM~\cite{yang2020combinational}&WACV2020&Heatmap&&$\surd$&$\surd$&&$\surd$&Yangseung/NL-CCAM\\
      ICMWSD~\cite{ren2020instance}&CVPR2020&SS&$\surd$&&$\surd$&&&\\
      \rowcolor{mygray}
      EIL~\cite{mai2020erasing}&CVPR2020&Heatmap&&$\surd$&$\surd$&&$\surd$&Wayne-Mai/EIL\\
      SLV~\cite{chen2020slv}&CVPR2020&SS&$\surd$&&$\surd$&&&\\
      \bottomrule
      \end{tabular}
\end{table*}
\addtolength{\tabcolsep}{3pt}

\subsection{Main Challenges}
\label{sec:challenge}

The main challenges of WSOD come from two aspects: localization accuracy and speed. For localization accuracy, it consists of discriminative region problem and multiple instances with the same category problem. For speed, it is an important characteristic of real applications. In TABLE~\ref{table08}, we summarize all typical WSOD methods and their contributions to these challenges.

\begin{figure}[t]
    \begin{center}
        \includegraphics[width=0.85\linewidth]{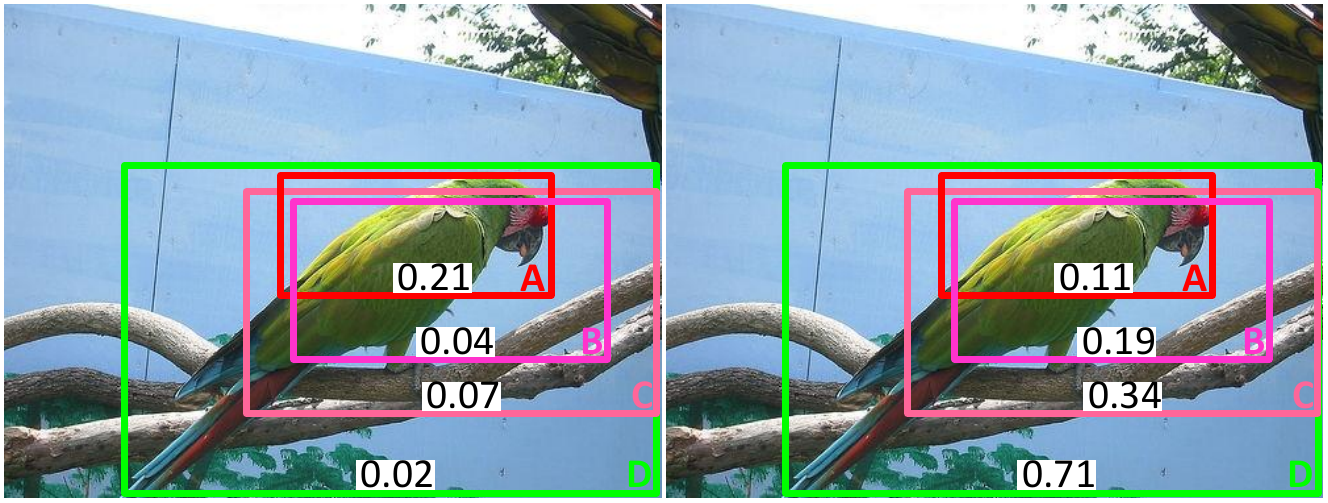}
    \end{center}
    \caption{Detection results between model without classifier refinement (left) and model with classifier refinement (right). The figure comes from~\cite{tang2017multiple}.} 
    \label{OICR_comparison}
\end{figure}

\textbf{Discriminative Region Problem.} It is that detectors~\cite{bilen2016weakly, zhou2016learning} tend to focus on the most discriminative parts of the object. During training, there may exist more than one proposal around an object, and the most discriminative part region of the object is likely to have the highest score (\eg, the region A is the most discriminative region in Fig.~\ref{OICR_comparison} (left) and it has a higher score than that of other regions). If the model selects positive proposals only based on scores, it is easy to focus on the most discriminative part of the object rather than the whole object extent.

\textbf{Multiple Instance Problem.} It denotes that detectors~\cite{bilen2016weakly, tang2017multiple} are difficult to accurately recognize multiple instances when there may exist several objects with the same category in an image. Although there are multiple instances with the same category in an image, these detectors~\cite{bilen2016weakly, tang2017multiple} only select the highest score proposal of each category as the positive proposal and ignores other possible instance proposals.


\textbf{Speed Problem.} At present, the speed bottleneck of the WSOD approaches is mainly concentrated in proposal generation. Selective search (SS)~\cite{uijlings2013selective} and Edge boxes (EB)~\cite{zitnick2014edge} that are widely used in WSOD are too time-consuming. 

\subsection{Basic WSOD Framework}
\label{sec:framework}

The basic framework of WSOD methods can be categorized into MIL-based networks and CAM-based networks according to the detection of multiple instances and a single instance.

\subsubsection{MIL-based Network}
\label{sec:mil-based}
\blue{When the detection network predicts multiple instances in an image, it is considered a Multiple Instance Learning (MIL) problem~\cite{dietterich1997solving}. Taking Fig.~\ref{fsod_vs_wsod} (b) for example, an image is interpreted as a bag of proposals in the MIL problem. If the image is labeled cat, it means that at least one of the proposals tightly contains the cat instance. Otherwise, all of the regions do not contain the cat instance (likewise for dogs). The MIL-based network is based on the structure of WSDDN~\cite{bilen2016weakly} that consists of three components: proposal generator, backbone, and detection head. }

\textbf{Proposal Generator.} Numerous proposal generators are usually used in MIL-based networks. 1) \textit{Selective search (SS)}~\cite{uijlings2013selective}: it leverages the advantages of both exhaustive search and segmentation to generate initial proposals. 2) \textit{Edge boxes (EB)}~\cite{zitnick2014edge}: it uses object edges to generate proposals and is widely used in many approaches~\cite{bilen2016weakly, li2016weakly, diba2017weakly, jie2017deep, zhang2018zigzag, tang2018weakly, shen2019cyclic}. 3) \textit{Sliding window (SW)}: it denotes that each point of the feature maps corresponds to one or more proposals in the relative position of the original image. And SW is faster than SS~\cite{uijlings2013selective} and EB~\cite{zitnick2014edge} in proposal generation. 

\textbf{Backbone.} \blue{With the development of CNNs and large scale datasets (\eg, ImageNet~\cite{russakovsky2015imagenet}), the pretrained AlexNet~\cite{krizhevsky2012imagenet}, VGG16~\cite{simonyan2014very}, GoogLeNet~\cite{szegedy2015going}, ResNet~\cite{he2016deep}, and SENet~\cite{hu2018squeeze} are prevailing feature representation networks for both classification and object detection.}

\textbf{Detection Head.} \blue{It includes a classification stream and a localization stream. The classification stream predicts class scores for each proposal, and the localization stream predicts every proposal's existing probability score for each class. Then the two scores are aggregated to predict the confidence scores of an image as a whole, which are used to inject image-level supervision in learning.}

\blue{Given an image, we first feed it into the proposal generator and backbone to generate proposals and feature maps, respectively. Then, the feature maps and proposals are forwarded into a spatial pyramid pooling (SPP)~\cite{he2015spatial} layer to generate fixed-size regions. Finally, these regions are fed into the detection head to classify and localize object instances.}

\subsubsection{CAM-based Network}
\label{sec:cam-based}
\blue{When the detection network only predicts a single instance in an image, it is considered an object localization problem. The CAM-based network is based on the structure of CAM~\cite{zhou2016learning}, which consists of three components: backbone, classifier, and class activation maps.}

\textbf{Backbone.} \blue{It is similar to that of the MIL-based network introduced in Section~\ref{sec:mil-based}.}

\textbf{Classifier.} \blue{It is designed to classify the classes of an image, which includes a global average pooling (GAP) layer and a fully connected layer. }

\textbf{Class Activation Maps.} \blue{It is responsible for locating object instances by using a simple segmentation technique. Because the class activation maps are produced by matrix multiplying the weight of the fully connected layer to the feature maps of the last convolutional layer, it spotlights the class-specific discriminative regions in every activation map. Therefore, it is easy to generate bounding boxes of every class by segmenting the activation map of the class. }

\blue{Given an image, we first feed it into the backbone to generate feature maps of this image. Then, the feature maps are forwarded into the classifier to classify the image's classes. Meanwhile, we matrix multiply the weight of the fully connected layer to the feature maps of the last convolutional layer to produce class activation maps. Finally, we segment the activation map of the highest probability class to yield bounding boxes for object localization.}

\subsubsection{Discussions}
\blue{In this section, we discuss the differences between MIL-based networks and CAM-based networks.}

\blue{Firstly, MIL-based network leverages SS~\cite{uijlings2013selective}, EB~\cite{zitnick2014edge} or SW to generate thousands of initial proposals, but CAM-based network segments the activation map to one proposal for each class. Therefore, MIL-based network is better than CAM-based network when detecting multiple instances with the same category in an image, but the training and inference speed of CAM-based network is faster than MIL-based.}

\blue{Secondly, because the size of the proposals produced by SS or EB is not consistent, MIL-based network leverages an SPP layer to generate fixed-size vectors followed by feeding these fixed-size vectors into the fully connected layers for later training. However, a CAM-based network leverages a GAP layer to generate a fixed-size vector on the feature maps. Then, it feeds the vector into a fully connected layer for classifying.}


\blue{Finally, Both MIL-based networks and CAM-based networks will face the discriminative region problem and multiple instance problem. In addition, MIL-based networks will face the training and test speed problem, since SS and EB are too time-consuming and yield plenty of initial proposals.}

\section{Milestones of WSOD}

\label{sec:milestones}
\begin{figure}[t]
    \begin{center}
        \includegraphics[width=0.85\linewidth]{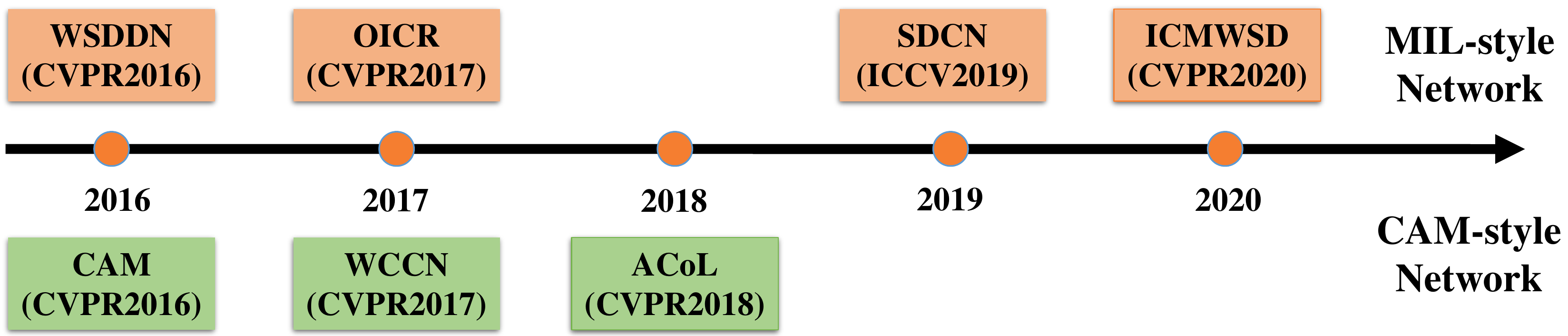}
    \end{center}
    \caption{The Milestones of WSOD since 2016.}
    \label{milestones}
\end{figure}

Since 2016, there are some landmark methods (cf. Fig.~\ref{milestones}) for the research of WSOD. In the following, we will briefly introduce these milestones.

\subsection{MIL-based Methods}
\label{sec:mil-based methods}

\textbf{WSDDN.} \blue{The biggest contribution of WSDDN~\cite{bilen2016weakly} is using two streams network, which aims to perform classification and localization respectively. WSDDN first uses a SPP~\cite{he2015spatial} on the top of the feature maps and generates a feature vector after two fully connected layer procedures. Next, the feature vector is fed into the classification stream and localization steam. Specifically, the classification stream is responsible for computing the class scores of each region, and the localization stream is designed to compute every region's existing probability for each class. Then, the matrix product of the class scores of each region and the existing probability for each class is considered as the final prediction scores. However, because of only accessing image-level labels in the training phase, the most discriminative part of the object will be paid more attention than the whole object instance in training. Due to the above limitation, WSDDN suffers from the discriminative region problem.}


\textbf{OICR.} \blue{To alleviate the discriminative region problem, OICR~\cite{tang2017multiple} uses WSDDN as its baseline and adds three instance classifier refinement procedures after the baseline. Every instance classifier refinement procedure, which consists of two fully connected layers, is designed to further predict the class scores for each proposal. Because the output of each instance classifier refinement procedure is the supervision of its latter refinement procedure, OICR can continue to learn so that larger area can have higher scores than WSDDN. Although the prediction of WSDDN may only focus on the discriminative part of the object, it will be refined after several instance classifier refinement procedures. }



\textbf{SDCN.} SDCN~\cite{li2019weakly} introduces a segmentation-detection collaborative mechanism. It consists of a detection branch and segmentation branch, which are responsible for detecting bounding boxes and generating segmentation masks respectively. In SDCN, the detection results will be converted to a heatmap by setting a classification score to all pixels within each proposal as the supervision mask of the segmentation branch. Meanwhile, the proposals of the highest overlap with the connected regions from the segmentation masks will be the pseudo ground-truth boxes of the detection branch. Both detection branch and segmentation branch are optimized alternatively and promoted each other, so SDCN achieves better detection performance than OICR.

\textbf{ICMWSD.} Different from SDCN which uses object detection with segmentation collaboration mechanism, ICMWSD~\cite{ren2020instance} addresses the problem of focusing on the most discriminative part of an object by leveraging context information. Firstly, ICMWSD obtains a dropped features by dropping the most discriminative parts. Then, maximizing the loss of the dropped features to force ICMWSD to look at the surrounding context regions.


\subsection{CAM-based Methods}
\textbf{CAM.} The biggest contribution of CAM~\cite{zhou2016learning} is using class activation maps to predict instances. CAM firstly leverages a GAP layer on the last convolutional feature maps to generate a feature vector. \blue{Then, the feature vector is fed into a classifier with a fully connected layer to generate prediction scores of an image.} Finally, CAM generates bounding boxes of each class by using a simple thresholding technique to segment the activation map of every class. \blue{However, class activation maps of CAM spotlight the regions that are the most discriminative parts of the object, so CAM also faces the discriminative region problem as WSDDN.}



\textbf{WCCN.} To alleviate the discriminative region problem, WCCN~\cite{diba2017weakly} proposes to use a cascaded network that has three cascade stages trained in an end-to-end pipeline. \blue{The first stage is the CAM~\cite{zhou2016learning} network that aims to generate class activation maps and initial proposals. The second stage is a segmentation network that uses the class activation maps to train object segmentation for refining object localization. The final stage is a MIL network that performs multiple instances of learning on proposals extracted in the second stage. Because the second and third stages refine object localization, WCCN alleviates the problem that tends to focus on the most discriminative part of the object.}

\textbf{ACoL.} To alleviate the discriminative region problem, ACoL~\cite{zhang2018adversarial} introduces two parallel-classifiers for object localization using adversarial complementary learning. Specifically, it first leverages the first classifier to localize the most discriminative regions. Then, ACoL uses the masked feature maps by masking the most discriminative regions discovered in the first classifier as the input feature maps of the second classifier. This forces the second classifier to select the next discriminative regions. Finally, ACoL fuses the class activation maps of both classifiers to generate bounding boxes of every class by segmenting the activation map of the highest probability class. 

\section{Specific Techniques for Discriminative Region Problem}
\label{sec:tec_discriminative}

In this section, we will introduce several advanced techniques for solving the discriminative region problem.



\subsection{Context Modeling}
The context of one region is external information of this region, which can be obtained by masking the region of the feature maps with special numbers (\eg, zero). There are two types of the strategy of using context modeling as follows.


\textbf{Strategy A.} It selects the regions that have a big gap between their scores and their contextual region's scores as positive proposals. For example, \blue{WSLPDA~\cite{li2016weakly} first replaces the pixel values within one proposal with zero to obtain the contextual region. Then, WSLPDA compares the scores of proposals and their contextual region. If the gap between the two scores is large, it indicates that the proposal is likely positive. ContextLocNet~\cite{kantorov2016contextlocnet} subtracts the localization score of one proposal from the localization score of the external rectangle region of the proposal. Then, the subtraction is considered as the final localization score of the proposal. Similar to WSLPDA and ContextLocNet, TS$^2$C~\cite{wei2018ts2c} selects a positive proposal by comparing the mean objectness scores of the pixels in one proposal and its surrounding region. But to alleviate the impact of background pixels in the surrounding region, TS$^2$C computes the mean objectness scores only using pixels with large confidence values in the surrounding region.}

\textbf{Strategy B.} It selects positive proposals by leveraging the loss of context regions. For example, \blue{OAILWSD~\cite{kosugi2019object} believes that a proposal not tightly covers the object instance if the loss of the context feature maps of this proposal tends to decrease. Thus, OAILWSD first leverages the context classification loss to label regions. Then, it selects the top-scoring regions whose context class probabilities are low as positive proposals. ICMWSD~\cite{ren2020instance} first drops the most discriminative parts of the feature maps to obtain contextual feature maps. Then, it maximizes the loss of the contextual feature maps to force it focuses on the context regions.}




\subsection{Self-training Algorithm}
In the self-training algorithm, the early prediction instances are then used in the detector for latter training as the pseudo ground-truth instances. There are two types of self-training algorithms: inter-stream and inter-epoch. In inter-stream self-training, the instances of each stream supervise its later stream. In inter-epoch self-training, the instances of each epoch supervise its later epoch. The key idea of self-training is that even if the early top-scoring proposals may only focus on the discriminative part of the object, they will be refined after several refinement procedures. 

\subsubsection{Inter-stream Self-training} 
OICR~\cite{tang2017multiple} expects B, C, and D can inherit the class score of A to correctly localize objects in Fig.~\ref{OICR_comparison} (right). So, OICR adds three refinement classifiers with two fully connected layers in WSDDN to address the issue shown in Fig.~\ref{OICR_comparison} (left). Specifically, the supervision of the first refinement classifier is the output of WSDDN. As for other refinement classifiers, the supervision of the current refinement classifier is the output of its previous refinement classifier. Inspired by OICR, WSOD2~\cite{zeng2019wsod2} consists of numerous classifiers. ICMWSD~\cite{ren2020instance} also inserts refinement streams in WSDDN, however, every refinement stream includes a classifier and a regressor. Besides, some approaches~\cite{zhang2018w2f, gao2019c, kosugi2019object, yang2019towards} use OICR as their baseline.

\subsubsection{Inter-epoch Self-training} 
\blue{Self-Taught-WS~\cite{jie2017deep} uses relative improvement (RI) of the scores of each proposal of two adjacent epochs as a criterion for selecting the positive sample. Specifically, it chooses the proposals of the previous epoch whose intersection over union (IoU) $\geq 0.5$ with the maximal RI proposal as the positive samples of the current epoch.}



\subsection{Cascaded Network}
The cascaded network includes several stages and the supervision of the current stage is the output of the previous stage. \blue{Such as WCCN~\cite{diba2017weakly} and TS$^2$C~\cite{wei2018ts2c} consist of three stages. The first stage is the CAM module that is to generate initial proposals using class activation maps. The intermediate stage is the object segmentation module that is designed to refine initial proposals. The final stage is a multiple instance learning module that is responsible for detecting accurate objects.}



\subsection{Bounding Box Regression}
Bounding box regression can improve object localization performance using instance-level annotations in the training phase, but the WSOD task only accesses image-level labels. To use bounding box regression for refining the initial proposals from SS~\cite{uijlings2013selective} or EB~\cite{zitnick2014edge}, some approaches propose to yield high-quality pseudo ground-truth boxes as the supervision of bounding box regression.

Now, numerous approaches~\cite{arun2019dissimilarity, zeng2019wsod2, yang2019towards, ren2020instance, chen2020slv} include at least one of the bounding box regressors. The supervision of the regressor is the output of previous classifiers. 



\subsection{Discriminative Region Removal}
From Fig.~\ref{OICR_comparison} (left), some researchers find that the highest score region only covers the most discriminative part of the object. To localize the whole object extent, masking the most discriminative part of the object is designed to force the detector to find the next discriminative region. 

TP-WSL~\cite{kim2017two} is a two-phase learning network that detects the next discriminative regions by masking the most discriminative region. In the first phase, it yields class activation maps followed by masking the most discriminative region using a threshold among the activation map of the highest probability class. In the second phase, it multiplies the masked activation map by the feature maps of the second network to refine the feature maps for detecting the next discriminative regions.

\blue{Different from TP-WSL that has two backbones, ACoL~\cite{zhang2018adversarial} consists of one shared backbone and two parallel-classifiers. The masked feature maps from the first classifier are fed into the second classifier to generate class activation maps. Finally, ACoL locates object instances in the fused activation maps by fusing the two-class activation maps of both classifiers. EIL~\cite{mai2020erasing} proposes to share the weights of the two parallel-classifiers of ACoL, and it only segments the activation map of the highest probability class from the unmasked branch to yields object proposals. Comparing C-MIDN~\cite{gao2019c} with ACoL, there are three differences. First, the detection network of C-MIDN is WSDDN~\cite{bilen2016weakly}, but the detection network of ACoL is CAM~\cite{zhou2016learning}. Second, C-MIDN does not compute the loss of high overlap with the first detection module's top-scoring proposal in the second branch, but ACoL masks the first detection module's top-scoring proposal's region with zero in the second branch. Finally, C-MIDN chooses the top-scoring proposals of the second detection module and the top-scoring proposals of the first detection module with low overlap with selected proposals as positive proposals, but ACoL yields positive proposals by segmenting the fused class activation maps.}

\subsection{Incorporating Low-level Features}
Low-level features usually retain richer object details, such as edges, corners, colors, pixels, and so on. We can obtain accurate object localization if making full use of these low-level features. For example, Grad-CAM~\cite{selvaraju2017grad} leverages high-resolution Guided Backpropagation~\cite{springenberg2014striving} that highlights the image's details to create both high-resolution and class-discriminative visualizations. WSOD2~\cite{zeng2019wsod2} first computes the score of a region proposal. Then, it selects the same region in low-level image features (\eg, superpixels) and computes the score of this region. Finally, the product of the two scores is the final score of the region proposal.




\subsection{Segmentation-detection Collaborative Mechanism}
Segmentation-detection collaborative mechanism includes a segmentation branch and a detection branch. The primary reasons for the collaborative mechanism are the following: 1) MIL (detection) can correctly distinguish an area as an object, but it is not good at detecting whether the area contains the entire object. 2) Segmentation can cover the entire object instance, but it cannot distinguish whether the area is a real object or not~\cite{li2019weakly}. So, some models leverage deep cooperation between detection and segmentation by supervising each other to achieve accurate localization.

WS-JDS~\cite{shen2019cyclic} first chooses the region proposals with top-scoring pixels generated by the semantic segmentation branch as the positive samples of the detection branch. Then, it sets the classification score to all pixels within each positive proposal of the detection branch as the supervision mask of the segmentation branch. \blue{Similar to WS-JDS, SDCN~\cite{li2019weakly} also combines the detection branch with the segmentation branch which is introduced in Section~\ref{sec:mil-based methods}.}


\subsection{Transforming WSOD to FSOD}
\label{sec:wsod_fsod}
Transforming WSOD to FSOD is another popular technique to achieve object detection using image-level labels, which is designed to train an FSOD model using the output of the WSOD model. The primary problem of transformation is to yield good pseudo ground-truth boxes from WSOD. There are several strategies to mine boxes as the pseudo ground-truth boxes. 1) \textit{top score}: numerous approaches~\cite{tang2017multiple, tang2018pcl, wei2018ts2c, shen2019cyclic, li2019weakly, gao2019c} select top score detection boxes of WSOD as the pseudo ground-truth boxes. 2) \textit{relative improvement (RI)}: ST-WSL~\cite{jie2017deep} selects the boxes with the maximal relative score improvement of two adjacent epochs as the pseudo ground-truth boxes. 3) \textit{mergence}: W2F~\cite{zhang2018w2f} merges several small boxes into a big candidate box and uses these merged boxes as the pseudo ground-truth boxes for later training. SLV~\cite{chen2020slv} first merges the scores of several boxes to the pixels and then generates bounding boxes of each class by using a simple thresholding technique to segment the map of every class.

In addition, there are several FSOD models that have been used as follows: Fast R-CNN~\cite{girshick2015fast}, Faster R-CNN~\cite{ren2015faster}, and SSD~\cite{liu2016ssd}. Numerous approaches~\cite{tang2017multiple, jie2017deep, tang2018pcl, wei2018ts2c, shen2019cyclic, li2019weakly, gao2019c, chen2020slv} use prediction boxes of WSOD as the pseudo ground-truth boxes to train Fast R-CNN. W2F~\cite{zhang2018w2f} uses prediction boxes of WSOD to train Faster R-CNN. GAL-fWSD~\cite{shen2018generative} uses prediction boxes of WSOD to train SSD. 



\subsection{Discussions}
\blue{In the previous sections, we individually introduce several techniques that are commonly used to improve the detection performance by detailed listing numerous approaches. In this section, we will compare and discuss these techniques. }

\blue{Firstly, context modeling and discriminative region removal are two similar techniques. Context modeling is to calculate the scores of the proposal and its context region respectively. Then it chooses the positive proposal derives from the two scores. On the other hand, the discriminative region removal is to directly erases top-scoring regions by setting zero value in the feature maps of the first branch followed by feeding the erased feature maps into the second branch.}

\blue{Secondly, the self-training algorithm usually co-occurs with bounding box regression. Bounding box regression is responsible for refining the initial proposals from SS~\cite{uijlings2013selective} or EB~\cite{zitnick2014edge}. And self-training algorithm is designed to refine the prediction result of the baseline. The core problem of both the self-training algorithm and bounding box regression is yielding good pseudo ground-truth boxes.}

\blue{Thirdly, the cascaded network and segmentation-detection collaborative mechanism are two similar techniques. They leverage the segmentation module to improve the performance of the object detection module. A cascaded network is a sequential structure that the previous module is responsible for training the latter module. However, the segmentation-detection collaborative mechanism is a circular structure that leverages deep cooperation between detection and segmentation supervising each other to achieve accurate localization.}

\blue{Finally, incorporating low-level features technique leverages the advantage of the high-resolution characteristics of low-level features to improve object localization. The key idea of transforming WSOD to FSOD technique is to make full use of the advantages of the network structure of the FSOD model (\eg, Fast R-CNN~\cite{girshick2015fast}).}


\section{Specific Techniques for Multiple Instance problem}
\label{sec:tec_multiple}

\blue{In this section, we will introduce how to make full use of the spatial relationship of proposals to solve multiple instance problem introduced in Section~\ref{sec:challenge}. Specifically, the two proposals that are far away from each other are likely to correspond to two object instances, while the two proposals with large overlap may correspond to the same object instance.}



\blue{ST-WSL~\cite{jie2017deep} leverages a graph to detect multiple instances with the same category in an image. It first chooses N top-scoring proposals of each positive class as the nodes of the graph. The edge between two nodes indicates a large overlap between them. Then it selects the greatest degree (number of connections to other nodes) nodes as positive proposals using Non-Maximum Suppression (NMS) algorithm ~\cite{neubeck2006efficient}. PCL~\cite{tang2018pcl} introduces the proposal cluster to replace the proposal bag that includes all of the proposals of each category. PCL assigns the same label and spatially adjacent proposals to the same proposal cluster. If proposals do not overlap each other, they will be assigned in different proposal clusters. Then, PCL selects the highest score proposal from each proposal cluster as the positive proposal. W2F~\cite{zhang2018w2f} iteratively merges the highly overlapping proposals with top-scoring into big proposals. Finally, these big proposals are considered positive proposals.}

\section{Specific Techniques for Speed Problem}

In this section, we will introduce several advanced techniques for solving the speed problem introduced in Section~\ref{sec:challenge}. The main reason for the slow speed is that the MIL-based method adopts SS~\cite{uijlings2013selective} or EB~\cite{zitnick2014edge} to generate a large number of initial proposals that most of which are negative.

The methods for improving speed can be broadly categorized into three groups: 1) \textit{Transformation-based}~\cite{zhang2018w2f, shen2018generative}: these approaches use their prediction boxes as the pseudo ground-truth boxes to train Faster R-CNN~\cite{ren2015faster} or SSD~\cite{liu2016ssd} and then use Faster R-CNN or SSD to infer images. 2) \textit{Sliding-window-based}~\cite{durand2016weldon, zhu2017soft}: these approaches use the sliding window technique to quickly generate proposals by walking through every point on the feature map. 3) \textit{Heatmap-based}~\cite{zhou2016learning, selvaraju2017grad, durand2017wildcat, kim2017two, zhang2018adversarial, zhang2018self, choe2019attention, xue2019danet, yang2020combinational, mai2020erasing}: these approaches segment the heatmap using a threshold to generate proposals to improve the speed of proposal generation.

\section{Training Tricks}
\label{sec:tricks}
Besides the techniques in the previous chapter, training tricks \blue{without changing network structure} also can improve detection results. In this section, we will introduce several training tricks for improving object localization.


\subsection{Easy-to-hard Strategy}
Previous approaches~\cite{bilen2016weakly, zhou2016learning, durand2016weldon, kantorov2016contextlocnet, tang2017multiple, diba2017weakly, kim2017two} use all of the images at once without a training sequence to train the detection model. The easy-to-hard strategy denotes that the model is trained by using the images with progressively increasing difficulty. In this way, the model can gain better detection results. For example, ZLDN~\cite{zhang2018zigzag} first computes the difficulty scores of images. Then, all of the images are ranked in an ascending order based on the difficulty scores. Finally, ZLDN uses the images with increasing difficulty to progressively train themselves.


\subsection{Negative Evidence}
Negative evidence contains the low-scoring regions, activations, and activation maps. For example, WELDON~\cite{durand2016weldon} uses the classification scores of the $k$ top-scoring proposals and the $m$ low-scoring proposals to generate the classification scores of the image by simply summing. WILDCAT~\cite{durand2017wildcat} leverages the $k^+$ highest probability activations and $k^-$ lowest probability activations of the activation map to generate the prediction score. NL-CCAM~\cite{yang2020combinational} uses the lowest probability activation maps. Specifically, it first ranks all of the activation maps in a descending order based on the probability of every class. Then, it fuses these class activation maps using a specific combinational function into one map, which is segmented to predict object instances.

\subsection{Optimizing Smoothed Loss Functions}
If the loss function of the model is non-convex, it tends to fall into sub-optimal and falsely localizes object parts while missing a full object extent during training~\cite{wan2019c}. So C-MIL~\cite{wan2019c} replaces the non-convex loss function with a series of smoothed loss functions to alleviate the problem that the model tends to get stuck into local minima. At the beginning of training, C-MIL first performs the image classification task. During the training process, the loss function of C-MIL is slowly transformed from the convex image classification loss to the non-convex object detection loss function.


\subsection{Discussions}
\blue{In the previous sections, we individually introduce several training tricks that are independent of the network structure. In this section, we will compare and discuss these tricks.}

\blue{Firstly, an easy-to-hard strategy is applied to the data phase, which is responsible for adjusting the order of the training images. Secondly, negative evidence acts on the training phase, which is designed to refine positive proposals or feature maps. Finally, optimizing smoothed loss functions act on the optimizing phase, which is responsible for avoiding suboptimal.}


\section{Datasets and Performance Evaluation}
\label{sec:datasets}
\subsection{Datasets}
Datasets play an important role in WSOD task. Most approaches of the WSOD use PASCAL VOC~\cite{everingham2010pascal}, MSCOCO~\cite{lin2014microsoft}, ILSVRC~\cite{russakovsky2015imagenet}, or CUB-200~\cite{welinder2010caltech} as training and test datasets.

\textbf{PASCAL VOC.} It includes 20 categories and tens of thousands of images with instance annotations. PASCAL VOC has several versions, such as PASCAL VOC 2007, 2010, and 2012. Specifically, PASCAL VOC 2007 consists of 2,501 training images, 2,510 validation images, and 4,092 test images. PASCAL VOC 2010 consists of 4,998 training images, 5,105 validation images, and 9,637 test images. PASCAL VOC 2012 consists of 5,717 training images, 5,823 validation images, and 10,991 test images.

\textbf{MSCOCO.} It is large-scale object detection, segmentation, and captioning dataset. MSCOCO has 80 object categories, 330K images ($>$200K labeled), 1.5 million object instances. In object detection, MSCOCO is as popular as PASCAL VOC datasets. Because MSCOCO has more images and categories than PASCAL VOC datasets, the difficulty of training on the MSCOCO dataset is higher than that of PASCAL VOC datasets.

\textbf{ILSVRC.} The ImageNet Large Scale Visual Recognition Challenge (ILSVRC) is a large-scale dataset. In ILSVRC, the model usually uses 200 fully labeled categories and 1,000 categories in object detection and object localization, respectively. ILSVRC has several versions, such as ILSVRC 2013, ILSVRC 2014, and ILSVRC 2016. Specifically, ILSVRC 2013 which is usually used in object detection has 12,125 images for training, 20,121 images for validation, and 40,152 images for testing. In addition, ILSVRC 2014 and 2016 inherit the ILSVRC 2012 dataset in object localization, which contains 1.2 million images of 1,000 categories in the training set. And ILSVRC 2012 dataset has 50,000 and 100,000 images with labels in the validation and test set, respectively.

\textbf{CUB-200.} Caltech-UCSD Birds 200 (CUB-200) contains 200 bird species which is a challenging image dataset. It focuses on the study of subordinate categorization. CUB-200-2011~\cite{wah2011caltech} is an extended version of CUB-200, which adds many images for each category and labels new part localization annotations. CUB-200-2011 contains 5,994 images in the training set and 5,794 images in the test set. 

\subsection{Evaluation Metrics}
In the state-of-the-art WSOD approaches, there are three standard evaluation metrics: mAP, CorLoc, and top error.

\textbf{mAP (mean Average Precision).} Average Precision (AP) is usually used in image classification and object detection. It consists of precision and recall. If $tp$ denotes the number of the correct prediction samples among all of the positive samples, $fp$ denotes the number of the wrong prediction samples among all of the positive samples, and $fn$ denotes the number of the wrong prediction samples among all of the negative samples, precision and recall can be computed as
\begin{equation}
    \begin{aligned}
    \text{recall} & = tp / (tp + fn), \\
    \text{precision} &= tp / (tp + fp),
    \end{aligned}
\end{equation}
where the correct prediction sample denotes IoU of the positive sample and its corresponding ground-truth box $\geq$ 0.5. Meanwhile, the IoU is defined as
\begin{equation}
    {\rm IoU}(b, b^g) =  area(b \cap b^g) / area(b \cup b^g),
\end{equation}
where $b$ denotes a prediction sample, $b^g$ denotes a corresponding ground-truth box, and $area$ denotes the region size of the intersection or union. The mAP is the mean of all of the class average precisions and is a final evaluation metric of performance on the test dataset.



\textbf{CorLoc (Correct Localization).} CorLoc denotes the percentage of images that exist at least one instance of the prediction boxes whose IoU $\geq$ 50\% with ground-truth boxes for every class in these images. CorLoc is a final evaluation metric of localization accuracy on the trainval dataset.

\textbf{Top Error.} \blue{Top error consists of Top-1 classification error (1-err cls), Top-5 classification error (5-err cls), Top-1 localization error (1-err loc), and Top-5 localization error (5-err loc). Specifically, Top-1 classification error is equal to $1.0 - cls_1$, where $cls_1$ denotes the accuracy of the highest prediction score (likewise for Top-1 localization error). Top-5 classification error is equal to $1.0 - cls_5$, where $cls_5$ denotes that it counts as correct if one of the five predictions with the highest score is correct (likewise for Top-5 localization error). Numerous approaches~\cite{zhou2016learning, zhang2018adversarial, zhang2018self, xue2019danet, yang2020combinational} use top error to evaluate the performance of the model.}

\subsection{Experimental Results}
\noindent\textbf{Results on Pascal VOC.} The results of state-of-the-art WSOD methods on datasets Pascal VOC 2007, 2010, and 2012 are shown in Table~\ref{table_voc}. The WSOD methods with ``+FR" denote that their initial predictions are fed into the Fast R-CNN~\cite{girshick2015fast} and serve as pseudo ground-truth bounding box annotations, \ie, these methods transform the WSOD into FSOD problems. From the results, we can observe the performance on all three Pascal VOC datasets have achieved unprecedented progress in recent few years (\eg, mAP 52.1\% in CVPR'20 vs. 29.1\% in CVPR'16 on Pascal VOC 2012). Meanwhile, comparing the methods and their counterparts with Fast R-CNN (\eg, OICR vs. OICR+FR), we can find the detection performance can be further improved by using this FSOD transforming strategy.


\textbf{Results on MSCOCO.} The results of state-of-the-art WSOD methods on dataset MSCOCO are shown in Table~\ref{table_coco}. We only report the AP metric, and the AP$_{50}$ denotes that the IoU threshold is equal to $0.5$. Similarly, the performance on MSCOCO also doubled in the last few years (\eg, AP$_{50}$ 11.5\% vs. 24.8\% in test set). Since MSCOCO contains more object categories than PASCAL VOC datasets, the results on MSCOCO are still far from satisfactory. However, the performance gains by transforming WSOD to FSOD is relative marginal (\eg, 0.7\% gains in AP for PCL model).



\textbf{Results on ILSVRC 2020 and CUB-200.} TABLE~\ref{table_ILSVRC_CUB} summaries the object localization performance of state-of-the-art WSOD methods on these two datasets. Compared to the PASCAL VOC and MSCOCO, quite few WSOD methods have evaluated their performance on these two benchmarks. From TABLE~\ref{table_ILSVRC_CUB}, we can find the performance gains are also significant (1-err cls 35.6\% vs. 27.7\% in ILSVRC 2012).

\addtolength{\tabcolsep}{-4.5pt}  
\begin{table}[t]
    \centering
    \caption{The summary of detection results (mAp (\%) and CorLoc (\%)) of state-of-the-art WSOD methods on Pascal VOC 2007, 2010, and 2012 datasets. The FR means Fast R-CNN~\cite{girshick2015fast}.}
    \label{table_voc}
        \begin{tabular}{lcccccc}
        \toprule
        \multirow{2}*{Approach}&
        \multicolumn{2}{c}{2007}&
        \multicolumn{2}{c}{2010}&
        \multicolumn{2}{c}{2012}\\
        
        \cmidrule (r){2-3} \cmidrule (r){4-5} \cmidrule (r){6-7}
        &mAP &CorLoc &mAP &CorLoc &mAP &CorLoc \\
        \midrule
        WSDDN~\cite{bilen2016weakly}$_{\text{CVPR2016}}$&39.3&58.0&36.2&59.7&-&-\\
        \rowcolor{mygray}
        WSLPDA~\cite{li2016weakly}$_{\text{CVPR2016}}$&39.5&52.4&30.7&-&29.1&-\\
        ContextLocNet~\cite{kantorov2016contextlocnet}$_{\text{ECCV2016}}$&36.3&55.1&-&-&35.3&54.8\\
        \rowcolor{mygray}
        OICR~\cite{tang2017multiple}$_{\text{CVPR2017}}$&42.0&61.2&-&-&38.2&63.5\\
        WCCN~\cite{diba2017weakly}$_{\text{CVPR2017}}$&42.8&56.9&39.5&-&37.9&-\\
        \rowcolor{mygray}
        ST-WSL~\cite{jie2017deep}$_{\text{CVPR2017}}$&41.7&56.1&-&-&39.0&58.8\\
        SPN~\cite{zhu2017soft}$_{\text{ICCV2017}}$&-&60.6&-&-&-&-\\
        \rowcolor{mygray}
        TST~\cite{shi2017weakly}$_{\text{ICCV2017}}$&34.5&60.8&-&-&-&-\\
        PCL~\cite{tang2018pcl}$_{\text{TPAMI2018}}$&45.8&63.0&-&-&41.6&65.0\\
        \rowcolor{mygray}
        GAL-fWSD~\cite{shen2018generative}$_{\text{CVPR2018}}$&47.0&68.1&\textbf{45.1}&\textbf{68.3}&43.1&67.2\\
        W2F~\cite{zhang2018w2f}$_{\text{CVPR2018}}$&52.4&70.3&-&-&47.8&69.4\\
        \rowcolor{mygray}
        ZLDN~\cite{zhang2018zigzag}$_{\text{CVPR2018}}$&47.6&61.2&-&-&42.9&61.5\\
        MELM~\cite{wan2018min}$_{\text{CVPR2018}}$&47.3&61.4&-&-&42.4&-\\
        \rowcolor{mygray}
        TS$^2$C~\cite{wei2018ts2c}$_{\text{ECCV2018}}$&44.3&61.0&-&-&40.0&64.4\\
        C-WSL~\cite{gao2018c}$_{\text{ECCV2018}}$&45.6&63.3&-&-&43.0&64.9\\
        \rowcolor{mygray}
        WSRPN~\cite{tang2018weakly}$_{\text{ECCV2018}}$&47.9&66.9&-&-&43.4&67.2\\
        C-MIL~\cite{wan2019c}$_{\text{CVPR2019}}$&40.7&59.5&-&-&46.7&67.4\\
        \rowcolor{mygray}
        WS-JDS~\cite{shen2019cyclic}$_{\text{CVPR2019}}$&45.6&64.5&39.9&63.1&39.1&63.5\\
        Pred NET~\cite{arun2019dissimilarity}$_{\text{CVPR2019}}$&53.6&\textbf{71.4}&-&-&49.5&70.2\\
        \rowcolor{mygray}
        WSOD2~\cite{zeng2019wsod2}$_{\text{ICCV2019}}$&53.6&69.5&-&-&47.2&\textbf{71.9}\\
        OAILWSD~\cite{kosugi2019object}$_{\text{ICCV2019}}$&47.6&66.7&-&-&43.4&66.7\\
        \rowcolor{mygray}
        TPWSD~\cite{yang2019towards}$_{\text{ICCV2019}}$&51.5&68.0&-&-&45.6&68.7\\
        SDCN~\cite{li2019weakly}$_{\text{ICCV2019}}$&50.2&68.6&-&-&43.5&67.9\\
        \rowcolor{mygray}
        C-MIDN~\cite{gao2019c}$_{\text{ICCV2019}}$&52.6&68.7&-&-&50.2&71.2\\
        ICMWSD~\cite{ren2020instance}$_{\text{CVPR2020}}$&\textbf{54.9}&68.8&-&-&\textbf{52.1}&70.9\\
        \rowcolor{mygray}
        SLV~\cite{chen2020slv}$_{\text{CVPR2020}}$ &53.5&71.0&-&-&49.2&69.2\\
        \hline
        OICR~\cite{tang2017multiple}+FR$_{\text{CVPR2017}}$ &47.0&64.3&-&-&42.5&65.6\\
        \rowcolor{mygray}
        PCL~\cite{tang2018pcl}+FR$_{\text{TPAMI2018}}$ &48.8&66.6&-&-&44.2&68.0\\
        MEFF~\cite{ge2018multi}+FR$_{\text{CVPR2018}}$ &51.2&-&-&-&-&-\\
        \rowcolor{mygray}
        C-WSL~\cite{gao2018c}+FR$_{\text{ECCV2018}}$ &47.8&65.6&-&-&45.4&66.9\\
        WSRPN~\cite{tang2018weakly}+FR$_{\text{ECCV2018}}$ &50.4&68.4&-&-&45.7&69.3\\   
        \rowcolor{mygray}   
        WS-JDS~\cite{shen2019cyclic}+FR$_{\text{CVPR2019}}$ &52.5&68.6&\textbf{45.7}&\textbf{68.1}&46.1&69.5\\
        SDCN~\cite{li2019weakly}+FR$_{\text{ICCV2019}}$ &53.7&\textbf{72.5}&-&-&46.7&69.5\\
        \rowcolor{mygray}
        C-MIDN~\cite{gao2019c}+FR$_{\text{ICCV2019}}$ &53.6&71.9&-&-&\textbf{50.3}&\textbf{73.3}\\
        SLV~\cite{chen2020slv}+FR$_{\text{CVPR2020}}$ &\textbf{53.9}&72.0&-&-&-&-\\
        \bottomrule
        \end{tabular}
\end{table}
\addtolength{\tabcolsep}{4.5pt} 

\addtolength{\tabcolsep}{-1.5pt} 
\begin{table}[t]
    \centering
    \caption{Detetion results on MSCOCO dataset comes from~\cite{ren2020instance}. These models use VGG16 as their convolutional neural network. There is no difference between AP and mAP under the MSCOCO context.}
    \label{table_coco}
    \scalebox{0.95}{
        \begin{tabular}{lccccc}
            \toprule
            \multirow{2}*{Approach}& 
            \multirow{2}*{Year}&
            \multicolumn{2}{c}{Val}&
            \multicolumn{2}{c}{Test}\\
            
            \cmidrule (r){3-4} \cmidrule (r){5-6}
            &&AP&AP$_{50}$&AP&AP$_{50}$\\
            \midrule
            WSDDN~\cite{bilen2016weakly}&CVPR2016&-&-&-&11.5\\
            \rowcolor{mygray}
            WCCN~\cite{diba2017weakly}&CVPR2017&-&-&-&12.3\\
            PCL~\cite{tang2018pcl}&TRAMI2018&8.5&19.4&-&-\\
            \rowcolor{mygray}
            C-MIDN~\cite{gao2019c}&ICCV2019&9.6&21.4&-&-\\
            WSOD2~\cite{zeng2019wsod2}&ICCV2019&10.8&22.7&-&-\\
            \rowcolor{mygray}
            ICMWSD~\cite{ren2020instance}&CVPR2020&\textbf{11.4}&\textbf{24.3}&\textbf{12.1}&\textbf{24.8}\\
            \hline
            Diba et al.~\cite{diba2017object}+SSD~\cite{liu2016ssd}&arXiv 2017&-&-&-&13.6\\ 
            \rowcolor{mygray}
            OICR~\cite{tang2017multiple}+Ens+FR~\cite{girshick2015fast}&CVPR2017&7.7&17.4&-&-\\
            MEFF~\cite{ge2018multi}+FR~\cite{girshick2015fast}&CVPR2018&8.9&19.3&-&-\\
            \rowcolor{mygray}
            PCL~\cite{tang2018pcl}+Ens.+FR~\cite{girshick2015fast}&TPAMI2018&9.2&19.6&-&-\\
            \bottomrule
        \end{tabular}
    }
\end{table}
\addtolength{\tabcolsep}{1.5pt}


\begin{table*}[t]
    \centering
    \caption{Object localization performance on ILSVRC 2012 and CUB-200-2011 datasets.}
    \label{table_ILSVRC_CUB}
        \begin{tabular}{lccccccccc}
        \toprule
        \multirow{2}*{Approach}& 
        \multirow{2}*{Year}&
        \multicolumn{4}{c}{ILSVRC 2012 (top error \%)}&
        \multicolumn{4}{c}{CUB-200-2011 (top error \%)}\\
        
        \cmidrule (r){3-6} \cmidrule (r){7-10}
        &&1-err cls&5-err cls&1-err loc&5-err loc&1-err cls&5-err cls&1-err loc&5-err loc\\
        \midrule
        CAM~\cite{zhou2016learning}&CVPR2016&35.6&13.9&57.78&45.26&-&-&-&-\\
        \rowcolor{mygray}
        ACoL~\cite{zhang2018adversarial}&CVPR2018&32.5&\textbf{12.0}&54.17&36.66&-&-&54.08&39.05\\
        SPG~\cite{zhang2018self}&ECCV2018&-&-&51.4&\textbf{35.05}&-&-&53.36&40.62\\
        \rowcolor{mygray}
        DANet~\cite{xue2019danet}&ICCV2019&32.5&\textbf{12.0}&54.17&40.57&\textbf{24.6}&\textbf{7.7}&47.48&38.04\\
        NL-CCAM~\cite{yang2020combinational}&WACV2020&\textbf{27.7}&-&\textbf{49.83}&39.31&26.6&-&47.6&\textbf{34.97}\\
        \rowcolor{mygray}
        EIL~\cite{mai2020erasing}&CVPR2020&29.73&-&53.19&-&25.23&-&\textbf{42.54}&-\\

        \bottomrule
        \end{tabular}
\end{table*}

\begin{table*}[t]
    \centering
    \caption{Some techniques and tricks for improving detection results and the approaches that utilize
    them. 1) Cont: Context modeling, 2) Self-t: Self-training algorithm, 3) Casc: Cascaded network, 4) BboxR: Bounding box regression, 5) DisRR: Discriminative region removal, 6) Low-l: Incorporating low-level features, 7) Seg-D: Segmentation-detection collaborative mechanism, 8) Trans: Transforming WSOD to FSOD, 9) E-t-h: Easy-to-hard strategy, 10) NegE: Negative evidence, 11) SmoL: Optimizing smoothed loss functions.}
    \label{table09}
        \begin{tabular}{lcccccccccccc}
        \toprule
        \multirow{2}*{Approach}&
        \multicolumn{8}{c}{Specific techniques for discriminative region problem} & 
        \multicolumn{3}{c}{Training tricks}\\
        
        \cmidrule (r){2-9} \cmidrule (r){10-12}
        &Cont&Self-t&Casc&BboxR&DisRR&Low-l&Seg-D&Trans&E-t-h&NegE&SmoL\\
        \midrule
        \rowcolor{mygray}
        WSDDN~\cite{bilen2016weakly}&&&&&&&&&&&\\
        CAM~\cite{zhou2016learning}&&&&&&&&&&&\\
        \rowcolor{mygray}
        WSLPDA~\cite{li2016weakly}&$\surd$&&&&&&&&&&\\
        WELDON~\cite{durand2016weldon}&&&&&&&&&&$\surd$&\\
        \rowcolor{mygray}
        ContextLocNet~\cite{kantorov2016contextlocnet}&$\surd$&&&&&&&&&&\\
        Grad-CAM~\cite{selvaraju2017grad}&&&&&&$\surd$&&&&&\\
        \rowcolor{mygray}
        OICR~\cite{tang2017multiple}&&$\surd$&&&&&&$\surd$&&&\\
        WCCN~\cite{diba2017weakly}&&&$\surd$&&&&&&&&\\
        \rowcolor{mygray}
        ST-WSL~\cite{jie2017deep}&&$\surd$&&&&&&$\surd$&&&\\
        WILDCAT~\cite{durand2017wildcat}&&&&&&&&&&$\surd$&\\
        \rowcolor{mygray}
        SPN~\cite{zhu2017soft}&&&&&&&&&&&\\
        TP-WSL~\cite{kim2017two}&&&&&$\surd$&&&&&&\\ 
        \rowcolor{mygray}
        PCL~\cite{tang2018pcl}&&$\surd$&&&&&&$\surd$&&&\\
        GAL-fWSD~\cite{shen2018generative}&&&&&&&&$\surd$&&&\\
        \rowcolor{mygray}
        W2F~\cite{zhang2018w2f}&&$\surd$&&&&&&$\surd$&&&\\
        ACoL~\cite{zhang2018adversarial}&&&&&$\surd$&&&&&&\\
        \rowcolor{mygray}
        ZLDN~\cite{zhang2018zigzag}&&&&&&&&&$\surd$&&\\
        TS$^2$C~\cite{wei2018ts2c}&$\surd$&&$\surd$&&&&&$\surd$&&&\\
        \rowcolor{mygray}
        SPG~\cite{zhang2018self}&&&&&&&&&&&\\
        WSRPN~\cite{tang2018weakly}&&&&&&&&&&&\\
        \rowcolor{mygray}
        C-MIL~\cite{wan2019c}&&&&&&&&&&&$\surd$\\
        WS-JDS~\cite{shen2019cyclic}&&&&&&&$\surd$&$\surd$&&&\\
        \rowcolor{mygray}
        ADL~\cite{choe2019attention}&&&&&&&&&&&\\
        Pred NET~\cite{arun2019dissimilarity}&&&&$\surd$&&&&$\surd$&&&\\
        \rowcolor{mygray}
        WSOD2~\cite{zeng2019wsod2}&&$\surd$&&$\surd$&&$\surd$&&&&&\\
        OAILWSD~\cite{kosugi2019object}&$\surd$&$\surd$&&&&&&&&&\\
        \rowcolor{mygray}
        TPWSD~\cite{yang2019towards}&&$\surd$&&$\surd$&&&&&&&\\
        SDCN~\cite{li2019weakly}&&&&&&&$\surd$&$\surd$&&&\\
        \rowcolor{mygray}
        C-MIDN~\cite{gao2019c}&&$\surd$&&&$\surd$&&&$\surd$&&&\\
        DANet~\cite{xue2019danet}&&&&&&&&&&&\\
        \rowcolor{mygray}
        NL-CCAM~\cite{yang2020combinational}&&&&&&&&&&$\surd$&\\
        ICMWSD~\cite{ren2020instance}&$\surd$&$\surd$&&$\surd$&&&&&&&\\
        \rowcolor{mygray}
        EIL~\cite{mai2020erasing}&&&&&$\surd$&&&&&&\\
        SLV~\cite{chen2020slv}&&$\surd$&&$\surd$&&&&$\surd$&&&\\
        \bottomrule
        \end{tabular}
\end{table*}

\section{Future Directions and Tasks}
\label{sec:directions}
Although we have summarized many advanced techniques and tricks for improving detection results, there are still several research directions that can be further explored.

\subsection{Model Directions}

\textbf{Better Initial Proposals.} The main proposal generators of the existing methods are selective search~\cite{uijlings2013selective}, edge boxes~\cite{zitnick2014edge}, heatmap, and sliding window. Selective search and edge boxes are too time-consuming and yield plenty of initial proposals that most of them are negative proposal. Segmenting heatmap tends to focus on the discriminative part of the object. The performance of the sliding window is strongly dependent on the size of objects. For example, if the size of the object instance is roughly fixed, then the sliding window works very well. Otherwise, it works badly. Because these generators have inherent disadvantages, we need to design a proposal generator that can yield fewer and more accurate initial proposals. The quality of the initial proposals directly affects the detection performance of the detector. So how to yield good initial proposals may be a new research direction.

\textbf{Better Positive Proposals.} Most WSOD methods select the proposals with the top score as positive proposals, which tend to focus on the most discriminative parts of the object rather than the whole object region. Because of the above problem, ST-WSL~\cite{jie2017deep} selects positive proposals according to the number of their surrounding proposals. And Self-Taught-WS~\cite{jie2017deep} selects positive proposals relying on the relative improvement (RI) of the scores of each proposal of two adjacent epochs. Besides, the key of self-training and cascaded network is to select accurate proposals as the pseudo ground-truth boxes for later training. Thus, how can we design a better algorithm that can accurately select positive proposals may be an important research direction.

\textbf{Lightweight Network.} Today's state-of-the-art object detectors~\cite{he2016deep, lin2017feature} leverage a very deep CNN to extract image feature maps and high-dimension fully connected layers to detect object instances that can achieve satisfactory detection performance. But the deep CNN and high-dimension fully connected layers rely on large memory and strong GPU computation power. Hence, a deep network is difficult to deploy on CPU devices (\eg, mobile phones). With the popularity of mobile devices, lightweight network with few parameters has received more and more attention from researchers, such as Light-Head R-CNN~\cite{li2017light}. Thus, designing a lightweight network in weakly supervised object detection may be a new research direction.

\subsection{Application Directions}

\textbf{Medical Imaging.} \blue{With the development of deep learning, it has evolved into cross-learning with multiple disciplines, especially the medical field. Because of lacking brain's Magnetic Resonance Imaging (MRI) and X-rays images with sufficient labels, weakly-supervised brain lesion detection~\cite{wu2019weakly, ji2019scribble} has received attention from researchers. The purpose of weakly-supervised brain lesion detection is to give the model the ability to accurately locate lesion region and classify lesion category that helps the doctor complete the diagnosis of the disease. Weakly-supervised lesion detection is not only applied in brain disease, but also other organ diseases, such as chest, abdomen, and pelvis. In addition to lesion detection, weakly-supervised learning is applied in disease prognosis~\cite{liu2019weakly}. During hospital visits, patients need to undergo a large number of tests to pinpoint their disease. These tests are generally presented to doctors and patients in the form of graphic reports. However, these numerous graphic reports lack correct labeling information. So, medical imaging may be another potential research direction in a weakly supervised setting.}

\textbf{3D Object Detection.} \blue{In recent years, with the continuous improvement of the accuracy of object detection of images~\cite{ren2015faster, liu2016ssd, redmon2016you, he2017mask, lin2017feature, law2018cornernet, ren2020salient, wu2021deformable, deng2021extended}, 3D object detection~\cite{chen2017multi, yang2019std, chen2020hierarchical, shi2020point} has received unprecedented attention. The purpose of 3D object detection is to detect object instances in the point cloud using 3D bounding boxes. Comparing with 2D object detection, 3D object detection tends to cost more computation and its supervision is more difficult to obtain and labor-intensive. Therefore, how to train light and accurate 3D detection models in the point cloud using simple labels may be a big challenge. Fortunately, weakly-supervised object detection is successfully applied in 2D object detection. According to the above analysis, we think that 3D weakly-supervised object detection that uses weak supervision(\eg, 2D bounding boxes and text labels) to train object detection models in the 3D scene may be a hot research direction.}


\textbf{Video Object Detection.} \blue{Video object detection~\cite{xiao2018video, deng2020single} is to classify and locate object instances in a piece of video. One of the solutions is to break the video into many frames and the detector achieves object detection in these frame images~\cite{deng2019object, chen2020memory}. However, the detector will face one big problem that the quality of these frame images has deteriorated. To improve the performance of video object detection, expanding the training dataset is a good approach. Unfortunately, tagging object location and category in videos is much more difficult than in 2D images. Therefore, training video object detection in the weakly-supervised setting is necessary.}



\section{Conclusions}
\label{sec:conclusion}
In this paper, we summarize plenty of the deep learning WSOD methods and give a lot of solutions to solve the above challenges. In summary, the main contents of this paper are listed as follows.
\begin{itemize}
    \item \blue{We analyze the background, and main challenges, and basic framework of WSOD. Furthermore, we introduce several landmark methods in detail.}
    \item For main challenges, we analyze almost all of the WSOD methods since 2016 and summarize numerous techniques and training tricks (cf. TABLE~\ref{table09}).
    \item We introduce currently popular datasets and important evaluation metrics in the WSOD task.
    \item We conclude and discuss valuable insights and guidelines for future progress in model and application directions.
\end{itemize}

\section*{Acknowledgment}
This work was supported by the National Natural Science Foundation of China (U19B2043, 61976185), Zhejiang Natural Science Foundation (LR19F020002, LZ17F020001), the Fundamental Research Funds for the Central Universities, Chinese Knowledge Center for Engineering Sciences and Technology.

\ifCLASSOPTIONcaptionsoff
  \newpage
\fi



\bibliographystyle{IEEEtran}
\bibliography{IEEEabrv,main_jrnl}
\end{document}